\documentclass[10pt,twocolumn,letterpaper]{article}

\usepackage{amsmath}
\usepackage{multirow}
\usepackage{booktabs}
\usepackage{pifont}
\usepackage[normalem]{ulem}
\usepackage{float}
\usepackage[accsupp]{axessibility}  

\DeclareUnicodeCharacter{0301}{\'{}}
\DeclareUnicodeCharacter{030C}{\v{}}

\def\arxivVersion{}

\ifdefined\arxivVersion
    \newcommand{\rev}[1]{{\color{magenta}#1}}
    \newcommand{\revexplain}{{\color{magenta}For the convenience of readers, the changes made during the rebuttal stage are highlighted in different colors.}}
\else
    \newcommand{\rev}[1]{#1}
    \newcommand{\revexplain}{}
\fi

\newcommand{\mypara}[1]{\noindent\textbf{#1.}}

\ifdefined\arxivVersion
    \usepackage[pagenumbers]{cvpr}   
\else
    \usepackage{cvpr}              
\fi

\definecolor{cvprblue}{rgb}{0.21,0.49,0.74}
\usepackage[pagebackref,breaklinks,colorlinks,allcolors=cvprblue]{hyperref}

\def\paperID{4346}
\def\confName{CVPR}
\def\confYear{2025}

\title{ArtFormer: Controllable Generation of Diverse 3D Articulated Objects}

\author{Jiayi Su$^{1,*,\dag}$\quad Youhe Feng$^{2,*}$\quad Zheng Li$^4$\quad Jinhua Song$^1$\quad Yangfan He$^{5,6}$\quad Botao Ren$^3$\quad Botian Xu$^{3,\ddag}$\\ 
$^1$Xiamen University Malaysia\quad $^2$Renmin University of China\quad $^3$Tsinghua University \\ $^4$Southern University of Science and Technology\quad $^5$University of Minnesota-Twin Cities \\ $^6$Henan RunTai Digital Technology Group Co., Ltd. \\
{$^*$ Equal Contribution.}\qquad
$^\dag$\tt\small{CST2209162@xmu.edu.my}\qquad
$^\ddag$\tt\small{btx0424@outlook.com}
}

\begin{document}

\newcommand{\figfirstpagefigure}{
    \vspace{-3.2em}
    \begin{center}
        \rev{\href{https://github.com/ShuYuMo2003/ArtFormer}     {github.com/ShuYuMo2003/ArtFormer}

        \revexplain
        }
        \vspace{1em}
        \captionsetup{type=figure}
        \includegraphics[width=1\textwidth]{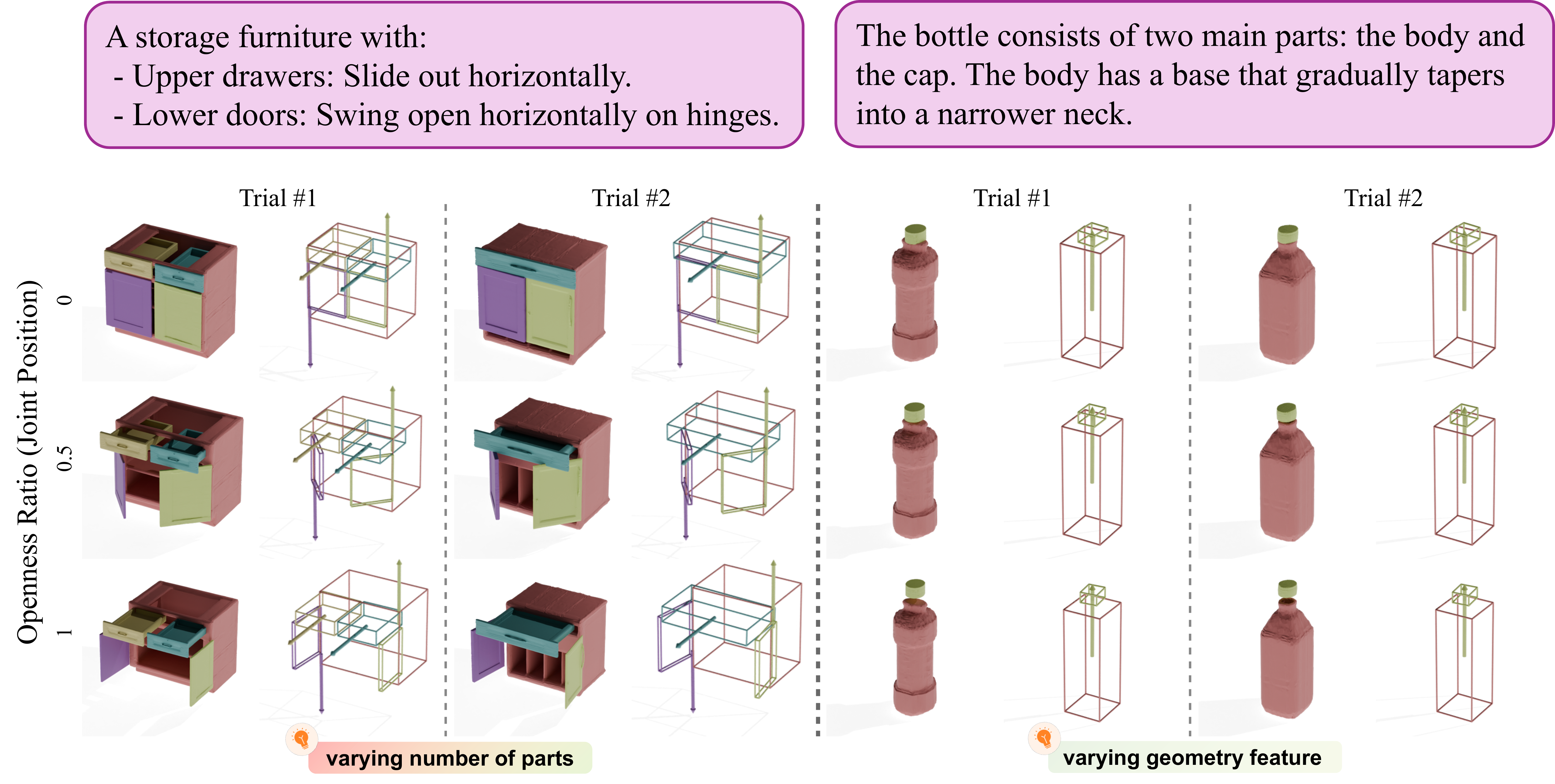}
        \vspace{-0.5em}
        \captionof{figure}{
            We present the \textbf{\emph{Art}}iculation Trans\textbf{\emph{Former}}, for high-quality generation articulated objects. This figure illustrates controlled generation across random trials based on text descriptions. Notably, it can generate a diverse range of objects with varying numbers of sub-parts and different geometry features. 
        }
        \label{fig:teaser}
    \end{center}
}

\twocolumn[{
    \maketitle
    \figfirstpagefigure
}]

\begin{abstract}


%
This paper presents a novel framework for modeling and conditional generation of 3D articulated objects. Troubled by flexibility-quality tradeoffs, existing methods are often limited to using predefined structures or retrieving shapes from static datasets. To address these challenges, we parameterize an articulated object as a tree of tokens and employ a transformer to generate both the object's high-level geometry code and its kinematic relations. Subsequently, each sub-part's geometry is further decoded using a signed-distance-function (SDF) shape prior, facilitating the synthesis of high-quality 3D shapes. Our approach enables the generation of diverse objects with high-quality geometry and varying number of parts. Comprehensive experiments on conditional generation from text descriptions demonstrate the effectiveness and flexibility of our method. 
\end{abstract}

\section{Introduction}
%


Articulated objects are defined as entities composed of multiple rigid sub-parts connected by various joints which allow the sub-parts to undergo constrained relative motion~\cite{liu2024survey}. Among others, man-made articulated objects constitute most everyday objects around us.


The perception~\cite{10.1145_1778765.1778795, jain2021screwnet, 10.1145_1531326.1531341,10.1111/cgf.12965} and reconstruction~\cite{mu2021asdf, wei2022nasam,heppert2023carto} of articulated objects have been extensively studied. However, research on generating articulated objects remains limited. On the one hand, to generate a multi-part articulated object, the model must simultaneously produce both the geometry of each sub-part and the kinematic relationships between them. Existing methods find it challenging to generate both modalities with high quality simultaneously. On the other hand, the complexity of articulated objects makes annotating them very costly, resulting in limited datasets for articulated objects.

Most relevant to this work are NAP \cite{nap}, CAGE \cite{cage}, and SINGAPO \cite{jiayi2024singapo}. They all support the conditional generation of 3D articulated objects but are limited to pre-defined graph structures. However, NAP has limited capability to adhere to the given condition while producing high-quality geometry. Meanwhile, aimed at controllability and quality, CAGE and SINGAPO do not actually generate the geometry but rather perform retrieval from datasets, restricting their ability to produce novel and diverse objects.


To achieve both diversity and usability, this paper proposes a novel framework, \textbf{\emph{Art}}iculation Trans\textbf{\emph{former}}, to generate high-quality and diverse articulated objects from text descriptions. We parameterize each {articulated} object with a tree structure. Each node corresponds to a sub-part, encompassing both its geometry and the kinematic relation (joint transform) relative to its parent node. Treating each node as a token, we utilize a transformer architecture to generate the sub-parts of the articulated object. Additionally, we introduce a tree position embedding in place of the ordinary position embeddings to better encode the tree structure from a sequence of tokens. Conditions (such as text descriptions and images) can be flexibly incorporated using cross-attention modules in the transformer layers.

However, simultaneously generating high-quality geometry and accurate joint parameters poses drastic challenges to both the model capacity and training pipeline. Instead of generating the geometries directly, we let the transformer output a compact latent code, which is then decoded by a Signed Distance Function (SDF) shape prior. The shape prior is trained on datasets with its latent space modeled by a diffusion model. This approach allows controllable sampling of sub-parts with varying geometry details.




In this paper, we primarily conduct experiments on text-guided generation of articulated objects, with a pre-trained text-encoder providing conditions to the transformer. Compared to prior works, our results demonstrate that we can generate a more diverse array of {articulated} objects that exhibit more precise kinematic features and high-quality geometry as well. Moreover, we also validate the flexibility of our framework with image-guided generation.

In summary, our main contributions are: \begin{enumerate}
        \item We present a novel framework for modeling and conditional generation of 3D articulated objects.

        \item A novel sampling and decoding recipe is designed to facilitate generation of shapes with diverse yet high-quality geometry.
        
        \item Through experiments on text- and image-conditioned generation, we validate the effectiveness and flexibility of our framework.



    \end{enumerate}
We believe our method could enable a range of future research and applications such as building Digital Cousins \cite{dai2024automatedcreationdigitalcousins} for scaling up robot learning.



\section{Related Work}


\subsection{Modeling 3D Articulated Objects.} 

Articulated objects are a specialized type of 3D object distinguished by their segmented, jointed structure, allowing for flexible movement and positioning of individual sub-parts. Modeling 3D articulated objects, an extension of 3D object modeling, involves the prediction~\cite{hauberg2011predicting, sturm2011probabilistic, morlans2023aograsp}, reconstruction, and generation of flexible, jointed structures.

Implicit neural representations have become a popular option recently \cite{wu20153dshapenetsdeeprepresentation,NIPS2017_ad972f10,park2019deepsdflearningcontinuoussigned,Mittal_2022_CVPR} due to their GPU memory efficiency and the ability to generate high-quality geometry. A convenient characteristic of implicit representations is that spatial transforms to the shape can be cast as rigid transforms to the input query points, making them a good choice for dealing with the kinematic relations in articulations. 

A-SDF \cite{mu2021asdf} is among the earliest explorers of using SDF to model articulated objects, but did not utilize the aforementioned property. More recently, NAP~\cite{nap} introduces the first 3D deep generative model for synthesizing articulated objects through a novel articulation tree/graph parameterization and the use of a DDPM~\cite{ho2020denoising}, enabling masked generation applications. Similarly, CAGE~\cite{cage} also employs a graph diffusion denoising model but with a primary aim of controllability. SINGAPO \cite{jiayi2024singapo} further extends controllable generation to single-image conditioning. \rev{MeshArt~\cite{gao2024meshartgeneratingarticulatedmeshes} utilizes domain-specific tokenizers to convert articulated objects into sequences and employs a transformer architecture for generation.}

However, CAGE and SINGAPO only generate abstractions of sub-parts, which are then used to retrieve similar assets from a dataset. Therefore, they can not produce objects with geometry features that are unseen in the dataset. This limitation is also common to methods that do not use SDF, such as URDFormer \cite{Chen2024URDFormerAP}, which predicts predefined URDF \cite{urdfref} primitives and meshes. A potential reason for such limitation is the difficulty of simultaneously modeling kinematic relations and geometry. Hyper-SDF \cite{Erko2023HyperDiffusionGI} and Diffusion-SDF \cite{diffusionsdf} propose methods to learn high-quality yet controllable priors of rigid SDFs. This work adopts a shape prior similar to Diffusion-SDF to ensure geometry quality.

\subsection{Generating Tree-structured Objects}

Generating tree-structured objects differs from conventional sequential generation, as each node can have multiple successors. Traditional approaches model graph distributions using Variational Autoencoders (VAEs)~\cite{kipf2016variational, simonovsky2018graphvae, zhang2019d}, Generative Adversarial Networks (GANs)~\cite{wang2018graphgan, de2018molgan, maziarka2020mol} and Denoising Diffusion Networks~\cite{hoogeboom2022equivariant, huang2022graphgdp, jo2022score, zhu2024mdiffusion, zhou2024unifying}. DiGress~\cite{vignac2022digress} and FreeGress~\cite{ninniri2024classifier}, for example, achieve state-of-the-art generation performance and can handle large, diverse molecular datasets. However, these methods are not tailored for tree-structured graphs and lack autoregressive generation, resulting in unreliable outputs for realistic acyclic, single-edge tree structures. To address this limitation, SceneHGN~\cite{scenehgn} introduces a recursive auto-encoder-based method that enables the hierarchical tree-structured generation of 3D indoor scenes. Similarly, \rev{GRAINS \cite{li2019grainsgenerativerecursiveautoencoders} achieves hierarchical indoor scenes generation using a specific recursive VAE.}
Shiv et al.~\cite{NEURIPS2019_6e091746} extend transformers to tree-structured data by {proposing} a novel tree-to-sequence mapping method. Peng et al.~\cite{NEURIPS2021_4e0223a8} advance this approach and enable Transformers to learn from both pairwise node paths and leaf-to-root paths by integrating tree path encoding into the attention module.

\begin{figure*}[!ht]
  \centering
  \includegraphics[width=\linewidth]{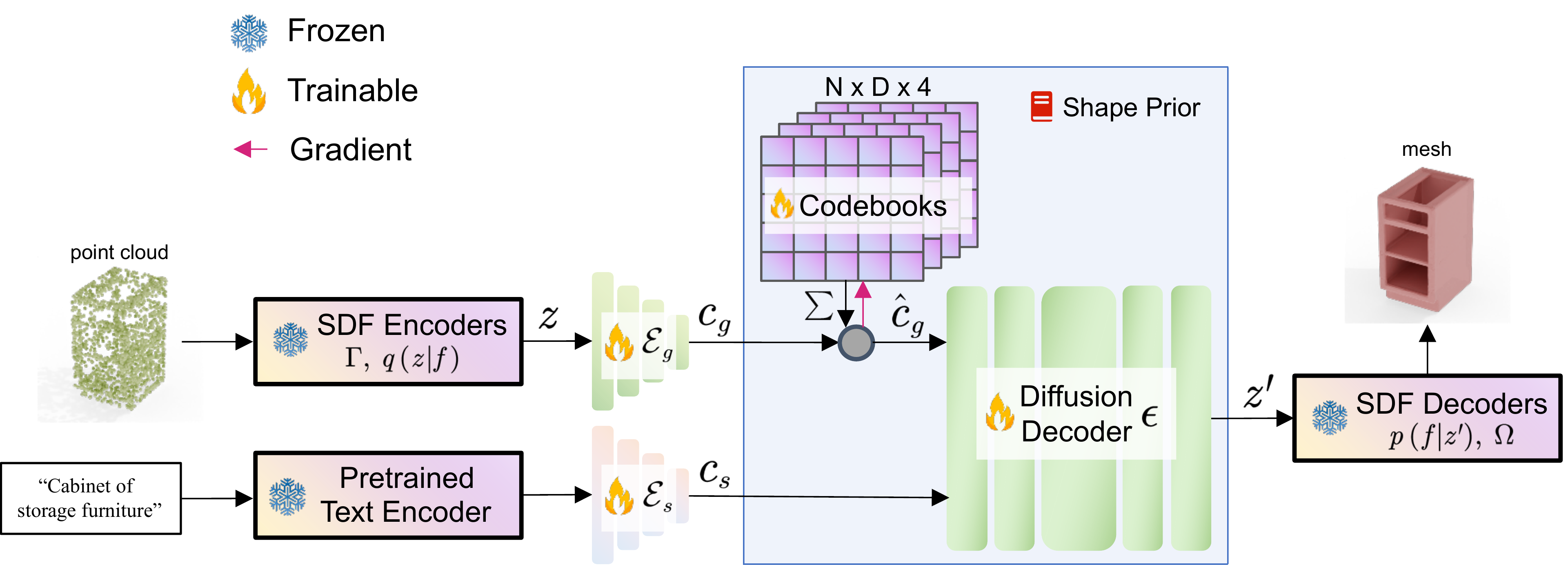}
  \caption{\textbf{Training Pipeline of Shape Prior} Mini encoder $\mathcal{E}_g$ compresses the geometry latent code $z$ into $c_g$, which is then processed by the embedding vectors of codebooks to form $\hat{c}_g$. $\hat{c}_g$ is the condition for diffusion decoder $\epsilon$. Each sub-part has a semantic label, such as `the lid of cup' or `handle of box'. These labels, encoded by the pre-trained text encoder, pass through mini encoder $\mathcal{E}_s$. The resultant vector $c_s$ is then passed into the diffusion shape prior directly.}
  \label{fig:shapeprior}
\end{figure*}

\section{Method}


\subsection{Articulation Parameterization}
\label{sec:parameterization}
Our parameterization process encodes an articulated object into a tree structure highly similar to the format used in URDF~\cite{urdfref} and MJCF \cite{todorov2012mujoco} files.
We consider each node (part) as a token that stores the geometry and kinematic relations of the corresponding sub-part of the articulated object. The attributes stored at each node are similar to those stored in the data parameterization of CAGE~\cite{cage}. Regarding the geometry information, for the $i$-th node, we identify the following $2$ attributes: \begin{itemize}
    \item \textbf{Bounding box, $b_i\in \mathbb{R}^6$}: For an articulated object, each sub-part is assigned an initial position, with its bounding box defining the maximum and minimum coordinates that the sub-part occupies along each axis in this initial state.
    \item \textbf{Geometry latent code, $z_i\in \mathbb{R}^{768}$}: We collect the point cloud of the sub-part. The point cloud is then processed through a series of encoders, converting it into a corresponding latent vector with dimension $768$ to represent the geometry of the $i$-th sub-part.
\end{itemize}

The kinematic parameters between $i$-th node and its parent node are represented by $2$ attributes: \begin{itemize}
    \item \textbf{Joint axis, $j_i \in \mathbb{R}^6$}: The joint axis includes an origin point and a unit direction vector. The $i$-th sub-part is capable of rotating around this axis or translating along it relative to its parent sub-part. The direction vector determines the positive direction for both rotational and translational movements.
    \item \textbf{Limit, $l_i \in \mathbb{R}^4$}: The attribute defines the permissible ranges for translational and rotational movements, setting the minimum and maximum extents of both translation and rotation relative to the initial position. If a sub-part is restricted from moving relative to its parent part, both the upper and lower bounds of these ranges are $0$.
\end{itemize}
For the $i$-th node in the tree, we store the aforementioned $4$ attributes, as well as the index of its parent node. Consequently, each node is represented by a token of dimension $D=6+768+6+4+1=785$. For all coordinates in each node, we utilize coordinates from the global coordinate system.

\subsection{Diverse and Controllable Shape Prior}
\label{sec:shapeprior}


As previously mentioned, simultaneous modeling and generating high-quality geometry and accurate kinematic relationships is challenging. Therefore, we first learn a \emph{shape prior} $p(z)$ of the geometry latent code using a method similar to Diffusion-SDF \cite{diffusionsdf}. 


\mypara{Shape Prior} An articulated object consists of multiple sub-parts. Given a sub-part sampled from the dataset, we encode its point cloud with a VAE encoder: $q(z|f)$ where $f=\Gamma(\rev{\text{point cloud}})\in \mathbb{R}^{3 \times 256 \times 64 \times 64}$ is an intermediate tri-plane feature obtained from a PointNet encoder $\Gamma$. A generalizable SDF network $\Omega(f, x)$ then predicts the part's SDF at query points $x\in R^3$ from decoded features $p(f|z)$. The training objective is:
\begin{equation}
\begin{split}
    L(q, p, \Gamma, \Omega) = & ||\Omega(\hat{f},x) - \text{SDF}(x)||_1 \\
    & + \beta D_\text{KL}\left(q(z|f)||\mathcal{N}(\mathbf{0}, \mathbf{I})\right).
\end{split}
\end{equation}
where $\hat{f}\sim p(f|q(z|f))$, $\text{SDF}(x)$ is the ground-truth signed-distance at point $x$ and $\beta$ balances the degree of regularization to a Gaussian prior. 

However, to subsequently enable guided or conditional generation of object sub-parts geometry, we train a conditional diffusion model $\epsilon (z_t, t, \hat{c}_g, c_s)$ on $p(z)$, where the geometry and semantic conditions are given by two corresponding encoders: $c_g=\mathcal{E}_g(z)$ and $c_s=\mathcal{E}_s(name)$. Before geometry condition $c_g$ is input into the diffusion model, it is processed into $\hat{c}_g$ using codebooks. The diffusion model is trained to denoise random latent $z_T\sim \mathcal{N}(0, \mathbf{I})$ into a meaningful $z_0\sim p(z)$, following the objective used in \cite{Ramesh2022HierarchicalTI}:
\begin{equation}
    L(\epsilon) = ||\epsilon(z_t, t, \hat{c}_g, c_s) - z_0||_2.
\end{equation} The aforementioned pipeline and structure can be referenced in \cref{fig:shapeprior}.
 
To process semantic information in text format (e.g., part name), we prepend a pre-trained text encoder \cite{t5} to $\mathcal{E}_s$. Note that after shape prior training, the two encoders $\mathcal{E}$ are discarded, leaving only the codebooks and diffusion decoder as our final shape prior.

\mypara{Sampling Diverse Shapes} A particularly desirable capability is to generate parts with diverse geometry features given the semantic information. For example, we would like USB caps of different shapes and styles. To enable our model for such capability, we discretize the space of geometry code $c_g=\mathcal{E}_g(z)$ to allow for sampling. A geometry condition $c_g$ is chunked into 4 segments $(c_g^0, c_g^1, c_g^2, c_g^3)$, \rev{which is designed to enlarge the capacity of the latent space from $4N$ to $N^4$}. \rev{Then, these 4 segments are} used to retrieve $(\hat{c}_g^0, \hat{c}_g^1, \hat{c}_g^2, \hat{c}_g^3)$ from 4 different codebooks $M_\rev{t} \in \mathbb{R}^{N\times D}$ \rev{($t$ is the index of codebook)} using Gumbel-Softmax sampling:
\rev{\begin{equation}
    \mathcal{D}_{t,l}=-||m_l^{\rev{t}}-c_g^\rev{t}||_2
    \label{eq:def_D}
\end{equation}}
\begin{equation}
    \hat{c}_g^\rev{t} = \sum_{j=1}^{N} \limits{m_j^{\rev{t}}\cdot\operatorname{GS}\left(\{\mathcal{D}_{t,l}\}_{l=1}^{N}\right)_j}
\end{equation}
where \rev{$\mathcal{D}_{t,l}$ is the distance matrix and } $m_l^{\rev{t}}\in \mathbb{R}^D$ denotes the $l$-th out of $N$ embedding vector in the codebook $M_\rev{t}$. The Gumbel-Softmax operation is defined as:
\begin{equation}
    \operatorname{GS}\left(\{x_\rev{l}\}_{\rev{l=1}}^\rev{N}\right)_j = \frac{\exp{\left((x_j + g_j) / \tau\right)}}{\sum_{\rev{k}=1}^{\rev{N}}\limits{\exp{\left((x_\rev{k} + g_\rev{k}) / \tau\right)}}}
    \label{eq:gs_sp}
\end{equation}
where $g_1, \cdots, g_k$ are samples from ${Gumbel}(0, 1)$~\cite{gsoft}. The softmax temperature $\tau$ controls the diversity of shape prior, which we do not specifically tune in this work. Since $\operatorname{GS}$ sampling is differentiable, the model can still be trained end-to-end.


\rev{The diffusion model helps sample from the full continuous space of $z$, given $\hat{c}_g$ which is either quantised from $c_g$ (produced by $\mathcal{E}_g$, during shape prior training) or sample by logits $P_i \in \mathbb{R}^{4\times N}$ from codebook $M$ directly (after shape prior training). By leveraging the shape prior, the sampled $z$ is guaranteed to align well with the target distribution.}
Meanwhile, the stochasticity introduced by discrete sampling improves the generation diversity.

\begin{figure}[!ht]
    \centering
    \includegraphics[width=\linewidth]{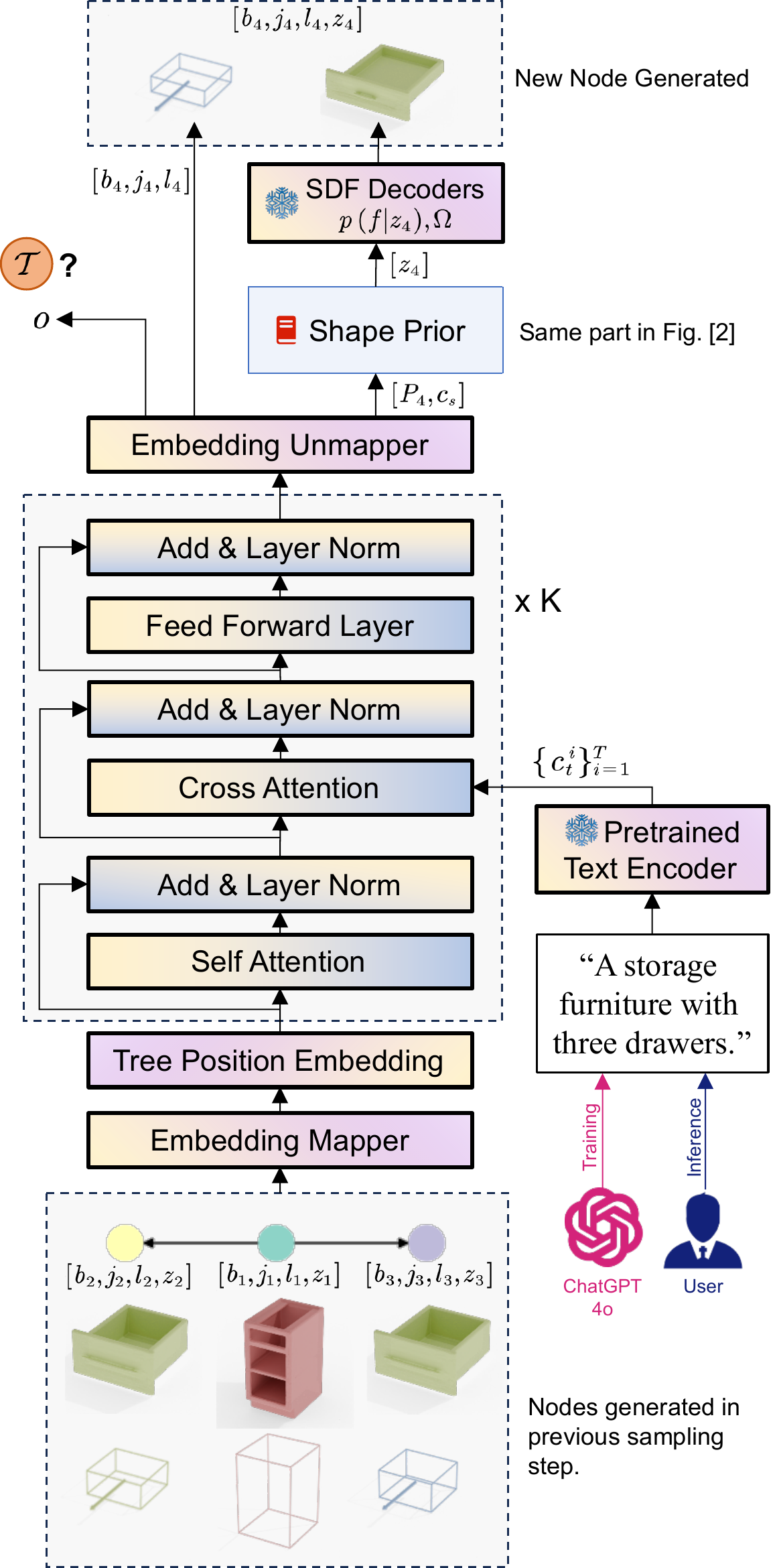}

   \caption{\textbf{Articulation Transformer}: In the tree structure, $i$-th node carries $4$ attributes: $b_i$, $j_i$, $l_i$ and $z_i$, which respectively represent the bounding box, joint axis, limit, and geometry latent code. $\hat{o}$ represents the logits indicating whether the current output token is a terminal token $\mathcal{T}$ (a special token).}
   \label{fig:artitrans}
\end{figure}

\subsection{Articulation Transformer}
\label{sec:transformer}

After the articulated object is parameterized into a tree structure, each node is treated as a token in the classical transformer architecture. The $i$-th token is composed of $[fa_i, b_i, z_i, j_i, l_i]$, where ${fa}_i$ is the parent index of the $i$-th node. In addition to the index of the parent node $fa_i$, the remaining attributes are concatenated and subsequently processed through an \rev{MLP Embedding Mapper}. An overview of the articulation transformer is illustrated in~\cref{fig:artitrans}.

\mypara{Tree Position Embedding} In order for the transformer to recognize the specific position of each token, we proposed a novel position encoding scheme specifically designed for tree structures \rev{building upon the works of \cite{treeposA,treeposB,treeposC}. }
We first calculate the absolute position encoding $a_i \rev{\in \mathbb{R}^{64}}$ for each $i$-th node:\begin{equation}
a_i=\operatorname{GRU}\left(\{\rev{\mathcal{A}_{p_k}}\}_{k=1}^{K}\right)
\end{equation} 
\rev{where $\mathcal{A}_i=[b_i, z_i, j_i, l_i]$, $K$ is the depth of $i$-th node and $\{p_k\}_{k=1}^K$ is the sequence of indices along the path on the tree from the root to $i$-th node.}
It pushes the tokens on the path from the root to a bi-directional GRU~\cite{treeposB, treeposC} to compress the information. We define the position embedding of the $i$-th node $p_i \rev{\in \mathbb{R}^{1024}}$ to represent the relative position:\begin{equation}
p_i=\operatorname{CAT}\left(\{a_{\rev{p_k}}\}_{k=K}^{\rev{1}} \right),
\end{equation} where $\operatorname{CAT}$ denotes concatenation. We employ truncation or padding with zeros to ensure that $p_i$ has a uniform length across all nodes.



\mypara{Conditioning} We primarily demonstrate conditional generation based on text descriptions. A text input is processed by a pre-trained text encoder~\cite{t5}, producing a sequence of conditioning tokens $\{c_t^i\}_{i=1}^T$, which are incorporated into the transformer through cross-attention layers. For training, we generate paired data using the following recipe: (1) sample an object from the dataset, (2) render images from different views using Blender \cite{blender}, and then (3) query ChatGPT-4o for text descriptions. For image conditioning, we can simply replace the text encoder with an image encoder and adopt a similar procedure. More details are provided in Supplementary Materials.


\mypara{Iterative Decoding}\label{sec:iterdec} 
Instead of predicting all parts at once (which assumes they are conditionally independent), we adopt an iterative decoding procedure to capture the inter-dependence between parts.

\begin{figure}[!ht]
    \centering
    \includegraphics[width=\linewidth]{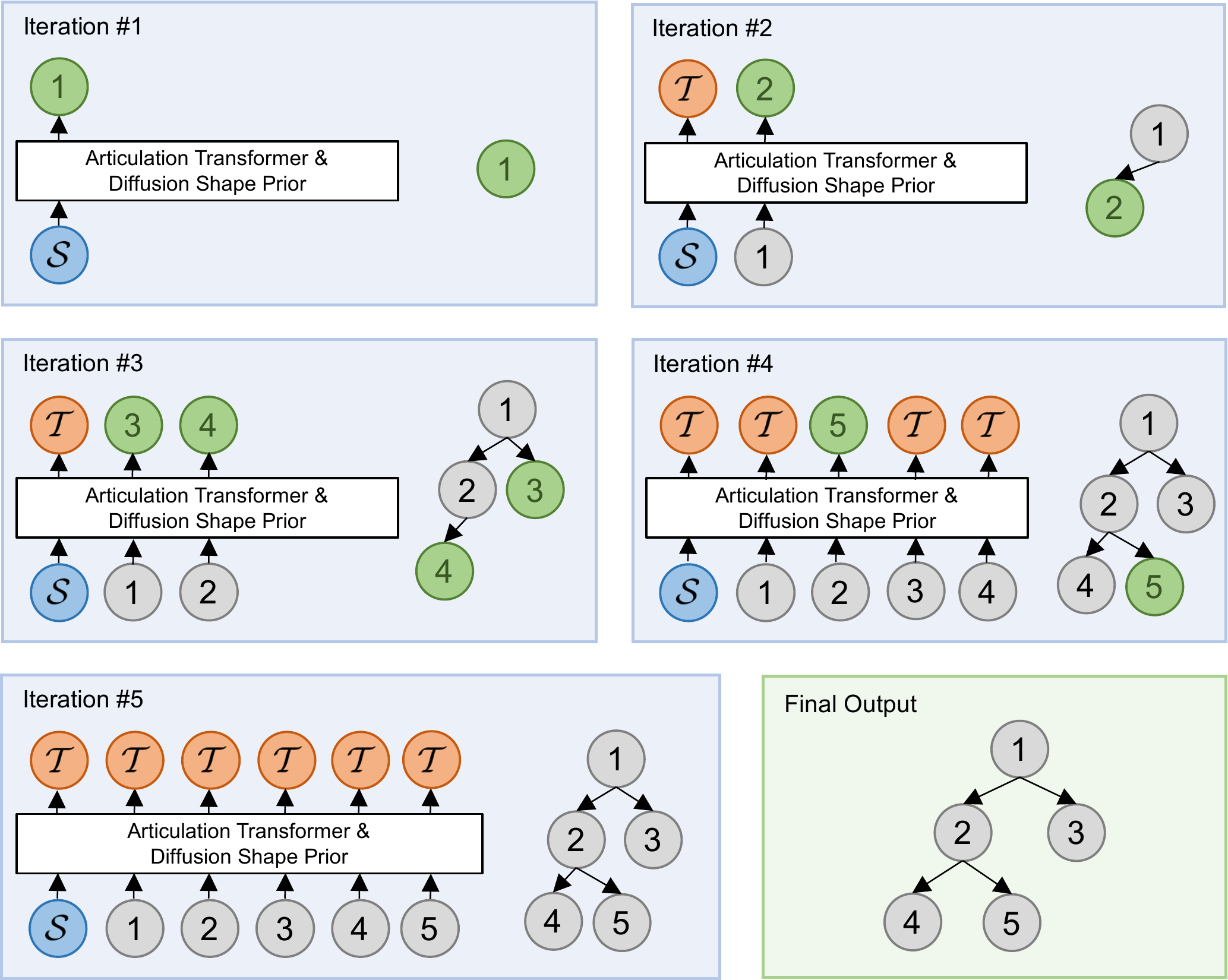}

   \caption{Each blue card represents a round in the predicting process. On each blue card, the left side shows the input given to the model and the expected output. The right side displays the tree structure of the articulated object formed after this prediction round, with green nodes indicating the nodes generated in this round. Orange nodes are terminal nodes.}
   \label{fig:pred_process}
\end{figure}

In each iteration, we input all previously generated nodes and predict a child for all of the input nodes, starting from a special start token $\mathcal{S}$, which conditions the generation of the root node. If no child nodes can be added to a current node, a special terminal token $\mathcal{T}$ is outputted. The self-attention layer ensures that the same child token is not repeatedly generated for any node across different prediction iterations. This process can be better understood through the illustration in \cref{fig:pred_process}.

The decoding procedure terminates when all output tokens are $\mathcal{T}$. We then \rev{project} the generated tokens \rev{back} to the format described in \cref{sec:parameterization}, i.e., the kinematic characteristics and relations $(b_i, j_i, l_i)$, and conditions $(P, c_s)_i$ for SDF decoding. To obtain the final object, cast joint transforms to rigid transforms to the input of SDF $\Omega$ and extract mesh using the Marching Cubes \cite{mcubes} algorithm. 

\mypara{Training Objective} \rev{The output of the articulation transformer are some tuples ($[o, b, j, l, c_s, P]_i$ for the $i$-th node).} Binary cross-entropy loss $L_o$ is employed to supervise $o$, which is the logits indicating whether this token is a terminal token. MSE loss $L_a$ is used to supervise the attributes $b, j, l, c_s$.


The training of the shape prior is conducted first. Upon completion, for each sub-part's geometry latent code $z$ in dataset, \rev{we compute the matrix $\mathcal{D}$ defined in ~\cref{eq:def_D}.}

$L_P$ denotes the loss function to supervise $P$: 
\begin{equation}
    L_P=\frac{1}{4}\sum_{i=0}^{3} D_{KL}\left( \mathcal{H}(P_{i, *}) || \mathcal{H}(\mathcal{D}_{i, *}) \right),
\end{equation} where $\mathcal{H}(P_{i, *})$ and $\mathcal{H}(\mathrm{D}_{i, *})$ are categorical distributions defined as: \begin{equation}
    \mathcal{H}(P_{i, *})(X = j) = \operatorname{softmax}(\{P_{i, l}\}_{l=1}^{N})_j, 
\end{equation}
\begin{equation}
    \mathcal{H}(\mathcal{D}_{i, *})(X = j) = \operatorname{softmax}(\{\mathcal{D}_{i, l}\}_{l=1}^{N})_j.
\end{equation}

The total loss function for \rev{Articulation Transformer} is defined as: \begin{align}
    L_{\text{trans}} = &\beta_{o}L_o + \beta_{P}L_P + L_a,
\end{align} where $\beta_{o}$ and $\beta_{P}$ are coefficients to balance the losses.
\section{Experiments}

\begin{table*}[t!]
\centering

\caption{Comparison of Generation Quality}
\begin{tabular}{@{}ccccccccc@{}}
\toprule
        & \multirow{2}{*}{\begin{tabular}[c]{@{}c@{}}Part Retrieval\end{tabular}} & \multirow{2}{*}{\begin{tabular}[c]{@{}c@{}}$\text{POR}\downarrow$\\ ${}_{\times 10^{-2}}$\end{tabular}} & \multicolumn{3}{c}{ID}   &  & \multicolumn{2}{c}{HS} \\ \cmidrule(lr){4-6} \cmidrule(l){8-9}  
        &                                                                            &                                                                  & MMD $\downarrow$    & COV $\uparrow$   & 1-NNA $\downarrow$ &  & AL $\uparrow$         & DS $\uparrow$       \\ \midrule
NAP-128 & \multirow{4}{*}{\ding{55}}                                                        & 0.805                                                            & 0.0710 & 0.3085 & 0.7021 &  & 0.105      & 0.13      \\
NAP-768 &                                                                            & 1.620                                                            & 0.0632 & 0.3723 & 0.6543 &  & 0.093      & 0.12      \\
\rev{Ours-NAPSP} &                                                                            & \rev{2.761}                                                           & \rev{0.0375} & \rev{0.4831} & \rev{0.8315} &  & \rev{-}      & \rev{-}      \\
Ours    &                                                                            & \textbf{0.709}                                                            & \textbf{0.0292} & \textbf{0.5213} & \textbf{0.5266} &  & \textbf{0.459}      & \textbf{0.67}      \\ \cmidrule(r){1-6}
CAGE    & \multirow{2}{*}{\ding{51}}                                                        & \textbf{0.251}                                                            & \textbf{0.0193} & 0.6064 & 0.5319 &  & 0.343      & 0.07      \\
Ours-PR &                                                                            & 0.556                                                            & 0.0214 & \textbf{0.6400} & \textbf{0.3950} &  & -         & -        \\ 
\bottomrule
\end{tabular}
\label{tab:comparison}
\end{table*}

\begin{figure*}[t!]
    \centering
    \includegraphics[width=1\linewidth]{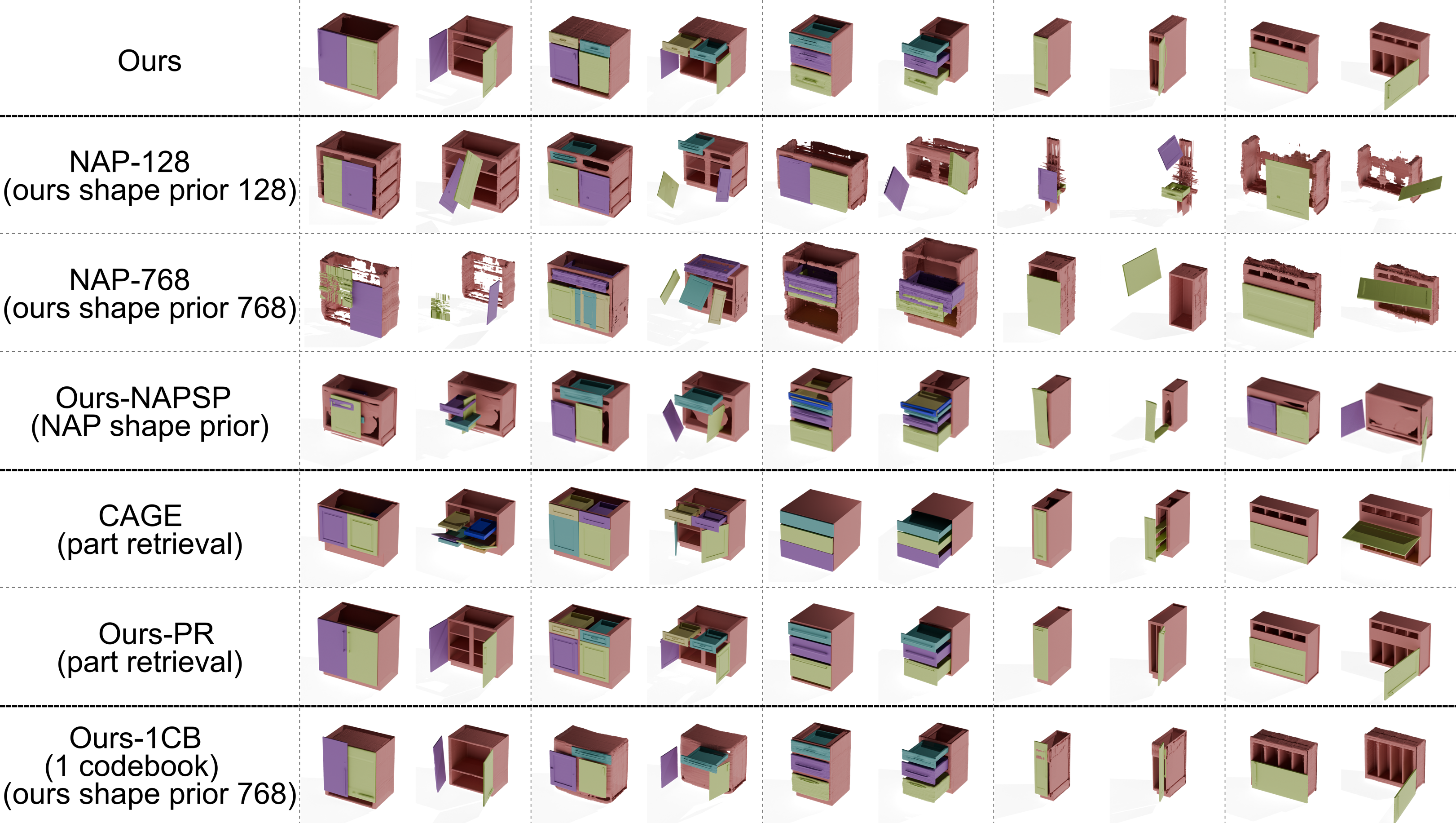}

   \caption{Qualitative comparison between \emph{ArtFormer} and baselines \rev{ (\texttt{Ours-1CB} will be discussed in~\cref{sec:ablation})}. Our method is capable of generating high-quality geometry and accurate joint relations.}
   \label{fig:comparison}
\end{figure*}

\subsection{Experimental Setup}

We train the shape prior on PartNet~\cite{Mo_2019_CVPR} and PartNet-Mobility~\cite{Xiang_2020_SAPIEN}. Although PartNet does not provide kinematic information, it still contributes to learning the geometry. The ArtFormer and other baseline models are trained exclusively on \rev{6 categories (Storage Furniture, Safe, Oven, USB, Bottle and Washer) in} PartNet-Mobility. For each articulated object in the dataset, we use Blender to create high-resolution thumbnails and employ ChatGPT-4o to generate corresponding descriptions, which are used for training the baselines. Detailed implementation steps can be found in Supplementary Material.

\subsection{Baselines}

Previous works, such as NAP~\cite{nap} and CAGE~\cite{cage}, differ from ours in several key aspects. NAP uses a simple shape prior with hidden dimensions that are not consistent with ours, while CAGE retrieves shapes from a dataset rather than generating them. To enable a fair comparison, we made modifications to these original models. A cross-attention layer, with the same structure as ours, is added to enable them to process text instructions. The compared models are:



 
\begin{enumerate}
    \item \textbf{NAP-128}: The original NAP model, modified to use our shape prior, generating a 128-dimensional shape code consistent with the original work.
    \item \textbf{NAP-768}: Building on NAP-128, we increase the size of the shape code to 768 dimensions to align with our model.
    \item \textbf{CAGE}: The original CAGE model, modified to retrieve outputs based on our shape prior.
    \item \textbf{Ours}: Our proposed model and articulation parameterization,  the geometry is generated through shape prior, bypassing the part retrieval.
    \item \textbf{Ours-PR}: Building on our original model. We perform part retrieval after iterative decoding, as CAGE does, for a fair comparison.
    \item \rev{\textbf{Ours-NAPSP}: Building on our original model, where the geometry is generated by the shape prior of NAP.}
\end{enumerate}

\subsection{Metrics}


We adopt the Instantiation Distance (ID) from NAP to evaluate the kinematic relations and geometry. A smaller ID value between two articulated objects indicates greater similarity, and vice versa. The following metrics are defined: (1) Minimum Matching Distance (\textbf{MMD}) describes the minimum distance between corresponding matches of generated objects and ground truths. (2) Coverage (\textbf{COV}) represents the ratio of ground truth instances that have a corresponding match in the generated objects, reflecting the similarity between the two distributions. (3) Nearest Neighbor Accuracy (\textbf{1-NNA}) measures the mismatch rate between generated and ground truth objects after performing 1-NN clustering.


To examine whether the generated objects are physically plausible, we also propose Part Overlapping Ratio (\textbf{POR}) which assesses the degree of interpenetration between sub-parts. Let $E$ represent the articulated object. We define the interpenetration metric between any two sub-parts $P_1, P_2 \in E$ as the vIoU (volume Intersection over Union) of their corresponding 3D geometries: 
\begin{align}
I(P_1, P_2)=\frac{|\mathcal G_1\cap \mathcal G_2|}{|\mathcal G_1\cup \mathcal G_2|},
\end{align}
where $\mathcal G_1$ and $\mathcal G_2$ represent the geometries of the sub-parts, respectively. Given a set of joint states $\mathcal{J}$, we can calculate the mean interpenetration between every pair of parts, denoted as $\operatorname{MI}(E,\mathcal J)$. We uniformly sample $N_j=10$ joint states $\mathcal J_1,\mathcal J_2,...,\mathcal J_{N_j}$ within the limits and define the Part Overlapping Ratio as:
\begin{align}
\operatorname{POR}(E)=\dfrac{1}{N_j}\sum\limits_{i=1}^{N_j} \operatorname{MI}(E,\mathcal J_i).
\end{align}





A human study methodology (HS) is used to assess the alignment between generated objects and their text descriptions, as well as the diversity of the generated objects. $P=20$ participants are presented with objects generated from $M=20$ distinct descriptions, each by 4 baseline models. Participants select the object that best matches the description. In a separate task, they choose the most diverse group from 4 options, each containing objects generated from the same instruction by different baselines. This task is repeated $T=5$ times. The alignment score (\textbf{AL}) and diversity score (\textbf{DS}) are defined as the mean win rate. Further details are provided in Supplementary Material.

\subsection{Results}

\mypara{Generation Quality and Diversity}
We evaluate the generation quality of baselines, as shown in \cref{tab:comparison}. Since CAGE cannot directly generate geometry, comparisons are divided into two groups. The first includes NAP-128, NAP-768, and our model, which generate geometry features directly. The second approach consists of CAGE and ours-PR, which retrieve the suitable shape from the dataset to generate objects. Our model outperforms NAP on all metrics in the first group, producing more realistic articulated objects with less part interpenetration. In the second group, while CAGE achieves better MMD and POR, indicating superior object-level reconstruction, our model excels in COV and 1-NNA, capturing the overall distribution and generating more diverse objects. HS results in \cref{tab:comparison} suggest that our model produces greater diversity and aligns better with text instructions from an ordinary user's perspective.

\begin{figure}[!ht]
    \centering
    \includegraphics[width=0.7\linewidth]{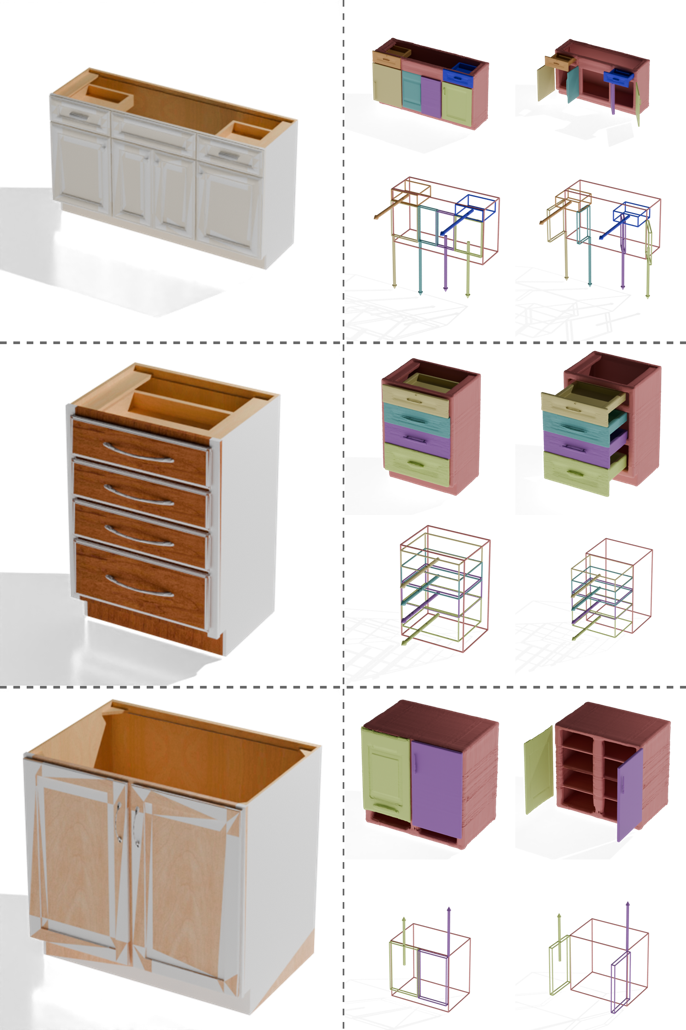}

   \caption{The figure displays 3 pairs of image condition inputs alongside the articulated object outputs produced by the model. Each pair contains a large image on the left as the input and generated articulated object on the right as the outputs.}
   \label{fig:image_cond}
\end{figure}

\mypara{Image Guided Generation} \label{sec:imagecond}
In our study, we replaced the original pre-trained text encoder~\cite{t5} with a pre-trained image encoder~\cite{blip2} to validate the flexibility of our method to support various conditioning modalities. We utilized Blender~\cite{blender} to render each object in the dataset as input images. The results of our experiment are shown in~\cref{fig:image_cond}. Our model is capable of generating high-quality articulated objects from a single image. This outcome further demonstrates its potential to scale to more complex and multi-modality settings.

\vspace{-5pt}

\subsection{Ablations}



\label{sec:ablation}
In our ablation study (\cref{tab:ablation}), we evaluated the impact of removing the tree position embedding (TPE) and shape prior (SP). Without the shape prior, the transformer must directly generate sub-part geometry, complicating feature generation, eliminating randomness, and significantly degrading all performance metrics. Removing the tree position embedding causes the model to lose positional information, increasing POR due to sub-part overlap during motion and reducing COV by impairing its ability to capture structural nuances and dataset distribution. \rev{We conduct the experiment with a single codebook (1 Codebook), and the performance drop is shown in \cref{fig:comparison} and \cref{tab:ablation}.}
\begin{table}[]
\centering

\caption{Ablation Studies with Reconstruction Quality}
\begin{tabular}{@{}ccccc@{}}
\toprule
        &  \multirow{2}{*}{\begin{tabular}[c]{@{}c@{}}$\text{POR}\downarrow$\\ ${}_{\times 10^{-2}}$\end{tabular}} & \multicolumn{3}{c}{ID}  \\ \cmidrule(lr){3-5}
        & & MMD$\downarrow$   & COV$\uparrow$   & 1-NNA$\downarrow$ \\ \midrule
Full    & {0.709} & 0.0292 & \textbf{0.5213} & 0.5266 \\
No TPE  & 1.170 & {0.0257} & 0.5000 & \textbf{0.5053} \\
No SP   & 2.502 & 0.0339 & 0.4574 & 0.7606 \\
\rev{1 Codebook} & \rev{\textbf{0.687}} & \rev{\textbf{0.0291}} & \rev{0.4948} &  \rev{0.5928} \\ 
\bottomrule
\end{tabular}
\label{tab:ablation}
\end{table}
\section{Conclusion}
We propose a novel method for modeling and generating 3D articulated objects, addressing limitations in diversity and usability. Representing articulated objects as a tree structure with nodes for rigid parts simplifies articulation parameterization, enabling part-level definition and generation. To ensure well-aligned yet diverse outputs, we develop a controllable shape prior and the Articulation Transformer, which captures articulation features effectively. A tree position embedding layer enhances part relationship modeling, supporting autoregressive generation. Our method achieves state-of-the-art performance, generating high-quality, diverse objects from text or image conditions. 


\mypara{Limitations and Future Work}
(1) The limited dataset restricts the application to a small range of object types with few \rev{sub-part}, preventing the full potential of the approach from being realized. \rev{Our model is expected to show more advantage when the number of parts is large ($>10$) and varies (which are rare in existing dataset), due to transformer's capability to handle long sequences of varying lengths.} Future work may explore this capability on \rev{a large scale}. 
(2) Multi-modal instructions beyond text and images have not yet been explored, such as point cloud or joint structure of expected articulated object. Investigating diverse instruction formats could greatly enhance flexibility and usability of our method in practical application. 
(3) \rev{Capturing joint quantitative details} in the text condition, such as rotation angles, is more challenging than joint type and geometry condition. Further research is needed to improve representation and learning of this data.
\rev{(4) It is observed that increasing the number of object categories during training leads to a decline in geometry reconstruction quality, likely due to the limited generalization of SDF models. Methods from GenSDF~\cite{gensdf} may enhance generalization, and we also recommend using modern 3D representations like 3D Gaussian Splatting~\cite{kerbl3Dgaussians} in future work.}

\mypara{Acknowledgements} This paper is a \textit{pure student work}, as all authors are either undergraduate or PhD students. 
Special thanks to \href{https://www.west-hpc.com/}{Ningxia Westcloud Computing Technology Co., Ltd.} for \href{https://www.damodel.com/}{DaModel’s} cloud services and student academic discounts, vital to this research.
Partial financial support was provided by \href{https://www.xmu.edu.my/}{Xiamen University Malaysia}. 

\clearpage

{
    \small
    \bibliographystyle{ieeenat_fullname}
    \bibliography{main}
}

\ifdefined\arxivVersion
    \clearpage
\setcounter{page}{1}
\maketitlesupplementary

\revexplain

\begin{figure*}[!htbp]
  \centering
  \includegraphics[width=\linewidth]{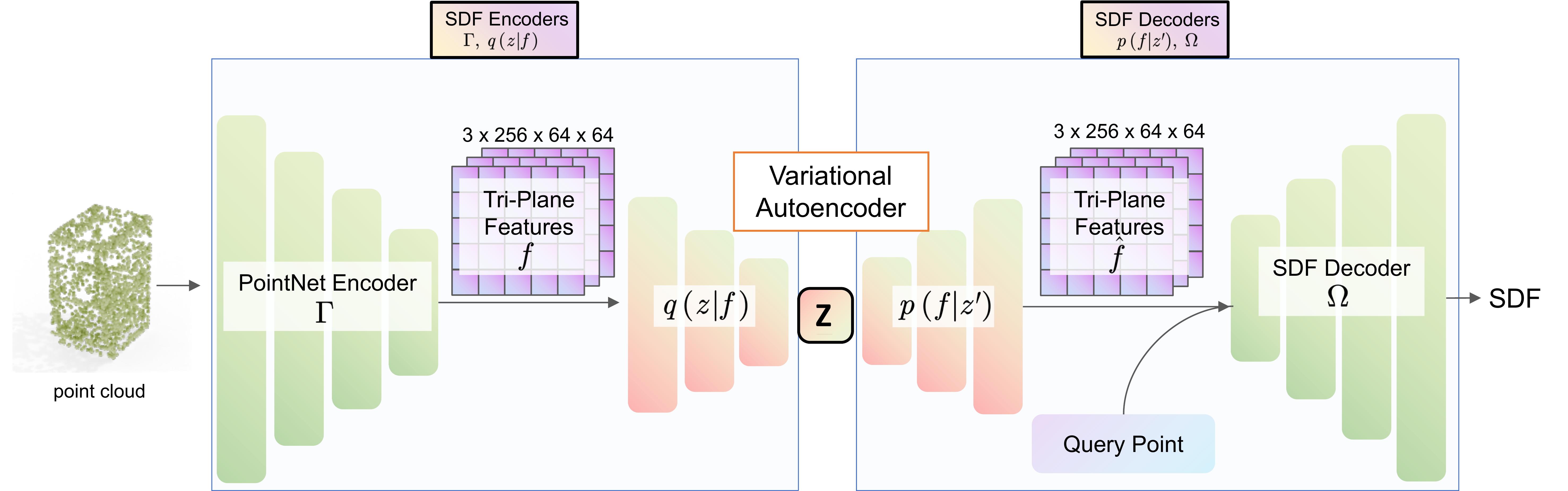}
  \caption{Training pipline for PointNet $\Gamma$, $q(z|f)$, $p(f|z)$ and SDF decoder $\Omega$. For brevity, we refer to $\Gamma$ and $q(z|f)$ collectively as SDF encoders. And, similarly, we refer to $p(f|z)$ and SDF decoder $\Omega$ as SDF decoders.}
  \label{fig:sdf}
\end{figure*}

\section{Implementation Details}

\subsection{SDF Model}
As we describe in~\cref{sec:shapeprior}, we firstly stack the PointNet $\Gamma$, $q(z|f)$, $p(f|z)$ and SDF Decoder $\Omega$, which is shown in~\cref{fig:sdf}. This stacked network is used to generate the latent code $z$ from the point cloud and decode the mesh from $z$. To strengthen the generalization of this network, we adopt a similar training method as~\cite{diffusionsdf}. During one training step of each sub-part, we randomly sample the point cloud which contains 4096 points, 16,000 query points $Q$, and compute the SDF value of $Q$. The training objective is $L(q, p, \Gamma, \Omega)$ as mentioned in~\cref{sec:shapeprior}.




\subsection{Training Details}
The training process employs the AdamW optimizer~\cite{Loshchilov2017DecoupledWD} with $\beta_1=0.9$ and $\beta_2=0.999$ for all of the models.

\mypara{SDF Model} We utilize articulated objects from two datasets, PartNet~\cite{Mo_2019_CVPR} and PartNet-Mobility~\cite{Xiang_2020_SAPIEN}, to train the SDF Model as displayed in ~\cref{fig:sdf}. The training on a single NVIDIA 4090 GPU with a batch size of 24 takes approximately 8 hours for 1.5k epochs.

\mypara{Diffusion Shape Prior} We use pretrained SDF model to generate the geometry latent code $z$ for each sub-part, which is used to train diffusion shape prior. The mini-encoders are implemented as 4-layer multilayer perceptron (MLP) networks. The dimensions of $c_g$ and $c_s$ are set to 64 and 32, respectively. Furthermore, the diffusion denoiser comprises $4$ blocks of normal transformers with self-attention layers.  The training on a single NVIDIA 4090 GPU with a batch size of 64 takes approximately 11 hours for 4k epochs.

\mypara{Articulation Transformer} The PartNet-Mobility dataset is exclusively used for training the Articulation Transformer. This network is composed of 8 transformer blocks, each with 8 attention heads and a token dimension of 1024. For the pre-trained text encoder, the encoder component of the T5 model~\cite{t5} is employed. The full training process takes 16 hours on a single NVIDIA 4090 GPU with a batch size of 128 for 17k epochs.

\subsection{Text Condition Generation Using GPT-4o}
\label{sec:gpt}
The training and testing of our model rely on text descriptions as conditions that highlight both kinematic and geometry features of articulated objects. 
Using prompt engineering, GPT-4o (\verb|gpt-4o-2024-08-06|) excels in image-to-text generation, producing detailed and precise descriptions for each object. The prompt we provide consists of two parts, $\mathcal{P}_{\text{base}}:\mathcal{P}_{\text{len},i}$, where~$:$ denote concatenation. The content of $\mathcal{P}_{\text{base}}$ is shown in \cref{fig:gpt4o_prompt}, and $\mathcal{P}_{\text{len},i}$, used to control the expected output length, is shown in \cref{fig:len_prompt}. For each object in the dataset, we supply GPT-4o with its corresponding snapshot and a series of text prompts $\{(\mathcal{P}_{\text{base}} : \mathcal{P}_{\text{len},i})\}_{i=0}^{3}$ sequentially, generating $4$ text conditions of varying lengths for the same object. 

In some cases, GPT-4o may fail to produce a valid description (e.g., returning \textit{"I'm sorry, I can't assist with that."}), with a failure rate of $26.10\%$. In the final dataset for text-guided generation, such failed descriptions are excluded. 


\begin{figure}[!htbp]
\centering
    \includegraphics[width=\linewidth]{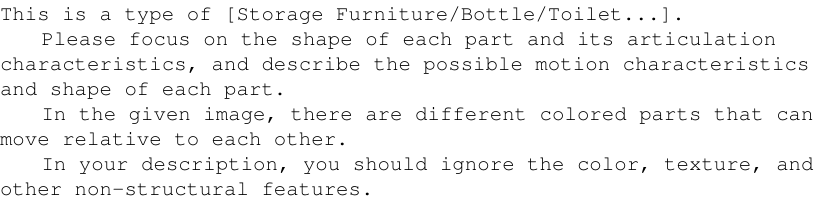}
    \caption{Prompt for GPT-4o to generate text description for objects.}
    \label{fig:gpt4o_prompt}
\end{figure}

\begin{figure}[!htbp]
\centering
    \includegraphics[width=\linewidth]{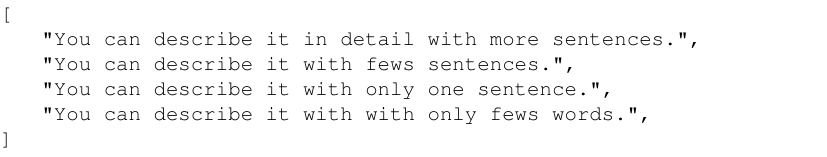}
    \caption{Prompt Used to Restrict the Length of Output.}
    \label{fig:len_prompt}
\end{figure}




   
\subsection{Human Study}
\label{sec:humanstudy}
We randomly selected 20 participants without prior knowledge of articulated object generation for the human study. Each participant completed the same questionnaire, divided into two sections for the alignment and diversity experiments, containing 20 and 5 questions, respectively. 

In the first section, participants evaluated four sets of images generated by different models from the same text instruction, with each set including three snapshots corresponding to openness ratios (linear interpolation between the predicted joint limits) of 0, 0.5, and 1. The text instruction used for generation was provided. The question is: \textit{select the set that best matched the described articulation characteristics and aligned with reality.}

In the second section, participants reviewed four snapshots with an openness ratio of 0, generated from the same instruction, repeated four times. The question is: \textit{select the set that shows the richest diversity while remaining consistent with reality.}










\section{Additional Experiments and Results}

\begin{figure}[!ht]
    \centering
    \includegraphics[width=1\linewidth]{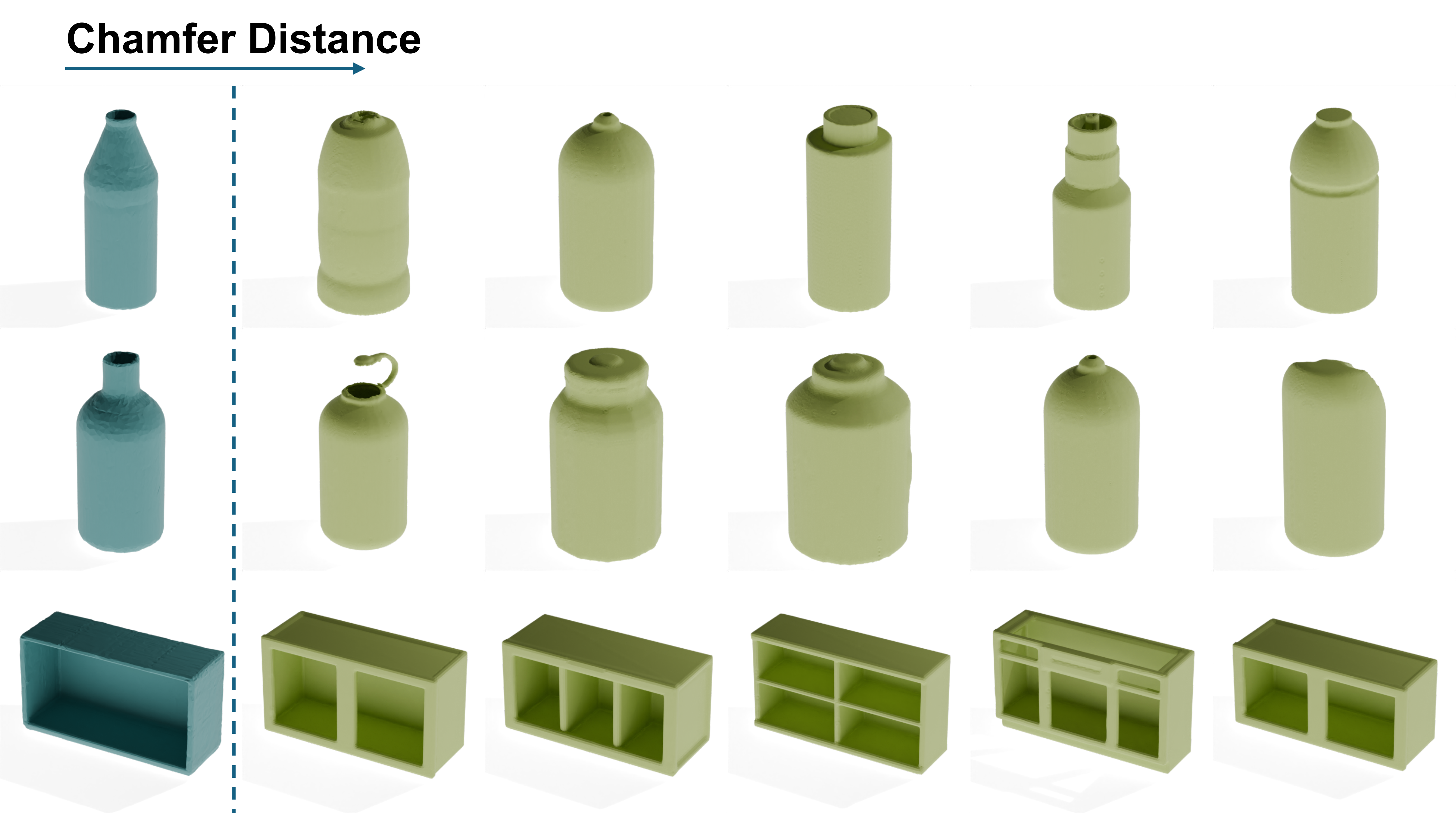}

   \caption{The first column (blue) shows shapes generated by our framework. The subsequent columns (green) are sub-parts retrieved from the training set, ordered according to increasing chamfer distance from the generated sub-parts.}
   \label{fig:cd}
\end{figure}

\rev{
\subsection{Novel Shape Generation} 
We conducted an experiment inspired by Diffusion-SDF to demonstrate that our shape prior, guided by an articulation transformer, can generate new geometry shapes that never appear in the dataset. We used our model to produce various objects and dissected them into sub-parts. Then, we calculated the Chamfer Distance between each sub-part and those in the training set and ranked them from nearest to farthest. The results, shown in \cref{fig:cd}, indicate that the sub-parts generated by our model are distinct from those in the training set, confirming the model's ability to create novel geometry shapes.
}

\subsection{Text Condition Attention}

\begin{figure*}[!htbp]
    \centering
    \includegraphics[width=0.5\linewidth]{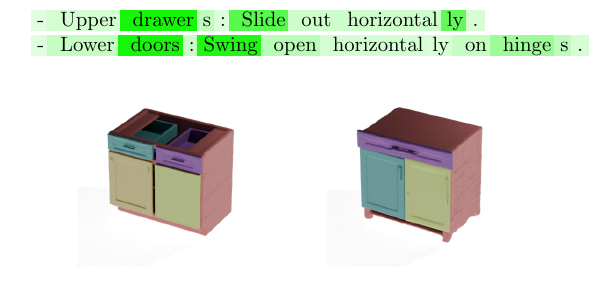}

   \caption{The text in the upper part of the figure represents the input text given to the model, where the intensity of the colors indicates the strength of attention; darker colors correspond to higher attention weights, while lighter colors indicate lower attention weights. The lower part of the figure displays the articulated objects generated by the model based on the input text.}
   \label{fig:text_atten}
\end{figure*}

To verify the weights of text in the transformer's cross-attention mechanism, we provided a brief description and calculated the average text token weight across each cross-attention layer. As shown in~\cref{fig:text_atten}, the words with higher attention weights describe the main sub-parts (`drawers' and `doors') of the object and the kinematic feature (`slide' and `swing') of these sub-parts. This confirms that our cross-attention mechanism effectively establishes relationships, allowing text conditioning to guide the generation of articulated objects.

\subsection{Editing of Articulated Objects}

\begin{figure*}[!htbp]
\centering
    \includegraphics[width=0.85\linewidth]{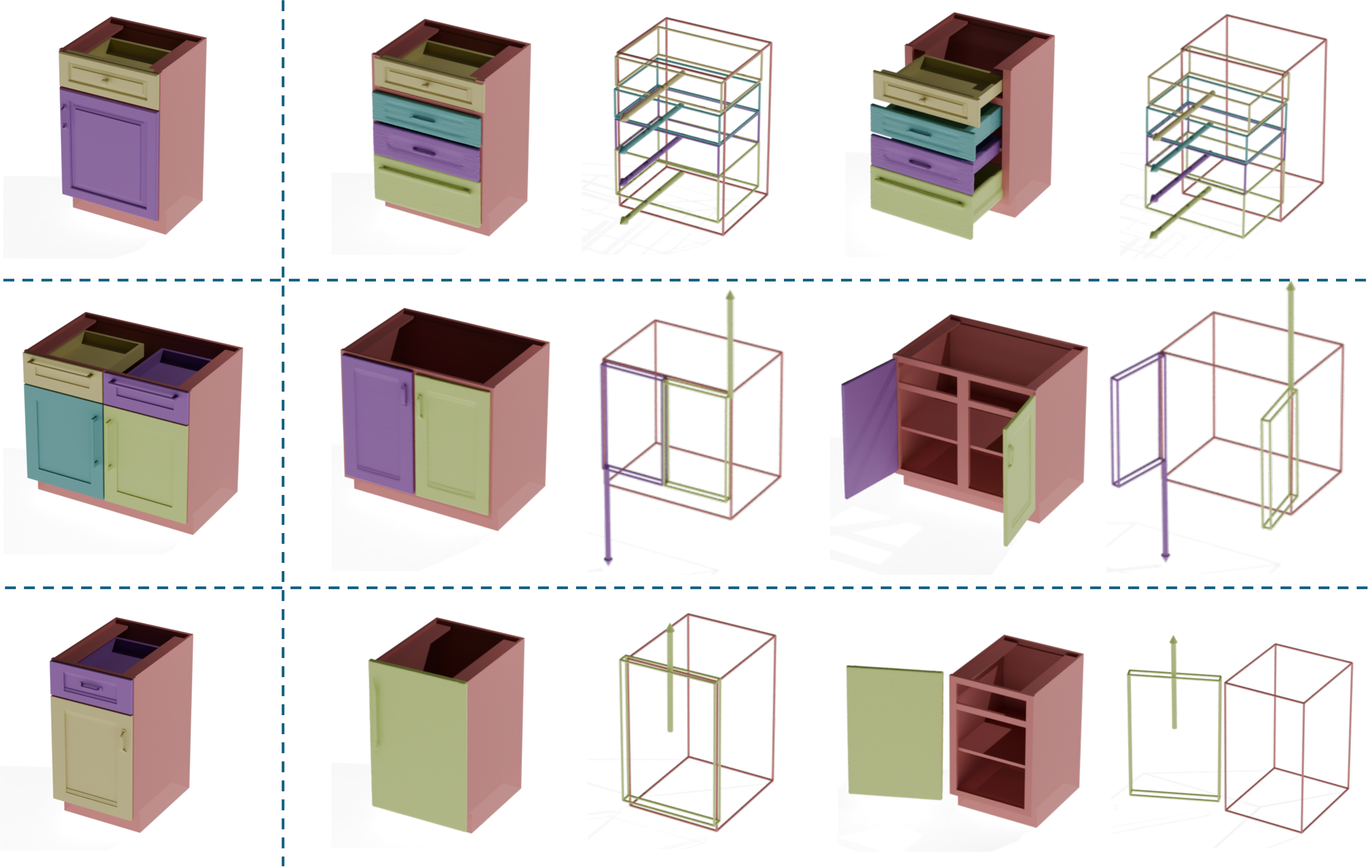}
    \caption{The figure illustrates the process of editing existing articulated objects using ArtFormer. (1) For the first object, the cabinet door (purple) is removed, and the following text condition is provided: \textit{This storage furniture consists of a rectangular frame with multiple horizontally aligned drawers that slide in and out on tracks}. The edited object is displayed on the right. (2) For the second object, the drawers (purple and yellow) and cabinet doors (green and blue) are removed. The condition is: \textit{This storage furniture consists of a rectangular base with two front panels that pivot on vertical hinges to open outward}. (3) For the third object, the drawer (purple) and cabinet door (yellow) are removed. The condition is: \textit{Rectangular frame: stationary base. Front panel: hinged door, pivots outward}. }
    \label{fig:editing}
\end{figure*}


To demonstrate the flexibility of autoregressive generation achieved through iterative decoding, we employed ArtFormer to edit existing articulated objects. In iterative decoding, each iteration generates a child node for each input node. This allows us to remove specific sub-parts from articulated objects and input desired text conditions, enabling the system to regenerate the missing sub-parts based on the provided text. In our experiment, we removed sub-parts from several objects in the training dataset and used ArtFormer to regenerate these incomplete parts based on alternative text instructions. The results are shown in \cref{fig:editing}.


\subsection{Text Guided Generation}

Additional visualization results are provided in~\cref{fig:text_cond_more} to illustrate the text guided generation for articulated objects using ArtFormer.

\begin{figure*}[!htbp]
\centering
    \includegraphics[width=0.85\linewidth]{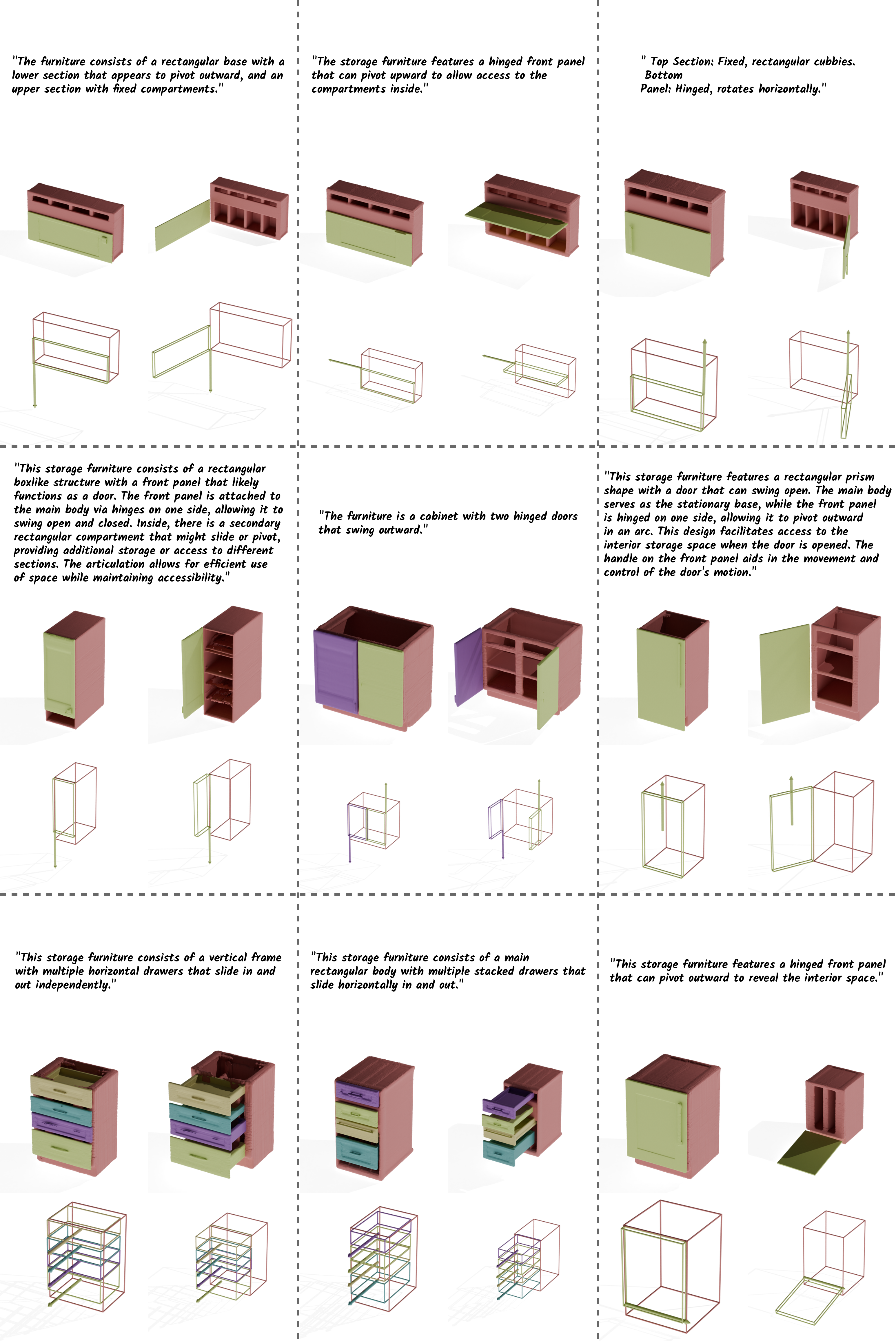}
    
    \caption{\textit{Continued on next page}}
\end{figure*}
\begin{figure*}[!htbp]
\ContinuedFloat
\centering
    \textit{Continued from previous page}
    
    \includegraphics[width=0.85\linewidth]{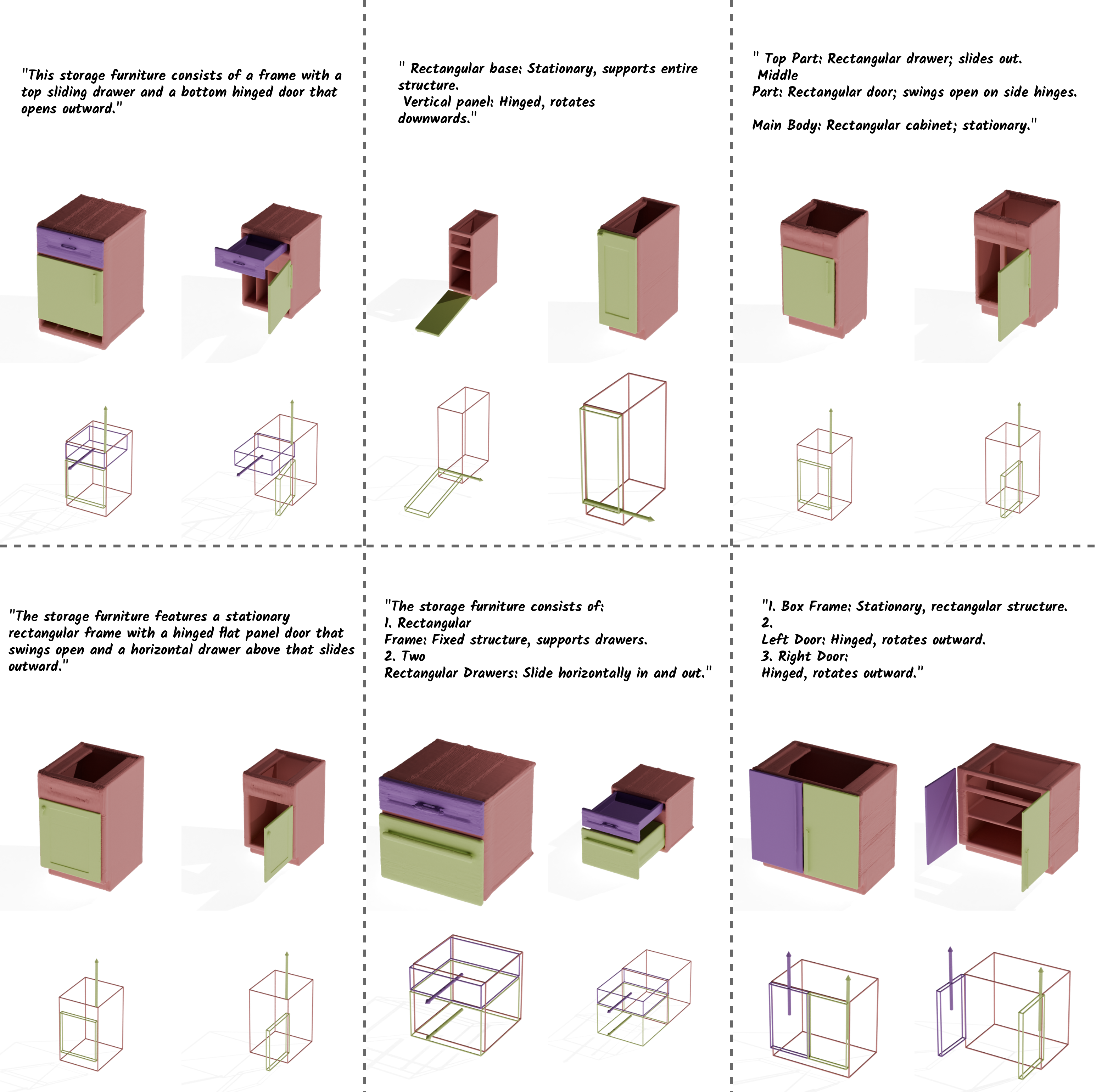}
    
    \caption{The figure presents 15 pairs of input text conditions and articulated objects generated by ArtFormer. For each pair, the text on the top serves as the input text condition, while the bottom side illustrates the articulated object output, showcasing the predicted motion relationship with the joint in both the fully closed and fully open states.}
    \label{fig:text_cond_more}
\end{figure*}

\rev{
A quantitative evaluation is conducted to assess text-guided generation alignment, complementing previous human study results. Inspired by Park et al.~\cite{park2021benchmark}, we adopt CLIP-R precision to measure the alignment between generated objects and instruction text. Snapshots are created for each generated object at an openness ratio of 0. The CLIP-R precision is then computed using these images and text instructions, leveraging the openai/clip-vit-large-patch14 model. The results for R=10 are presented in \cref{tab:clipr}.
}

\subsection{Image Guided Generation}

Preliminary experiments of ArtFormer's capability to generate articulated objects based on a single image, achieved by substituting the pretrained text encoder~\cite{t5} with a pretrained image encoder~\cite{blip2}, as discussed~\cref{sec:imagecond}. Rendered images of articulated objects from PartNet-Mobility~\cite{Xiang_2020_SAPIEN} using Blender, used as image conditions for generating articulated objects in the images with ArtFormer. The results are displayed in~\cref{fig:image_cond_more}. In addition, we employ real-world photographs as the image condition to generate articulated objects. The results are illustrated in~\cref{fig:real_image_cond}.


\begin{figure*}[!htbp]
\centering
    \includegraphics[width=0.85\linewidth]{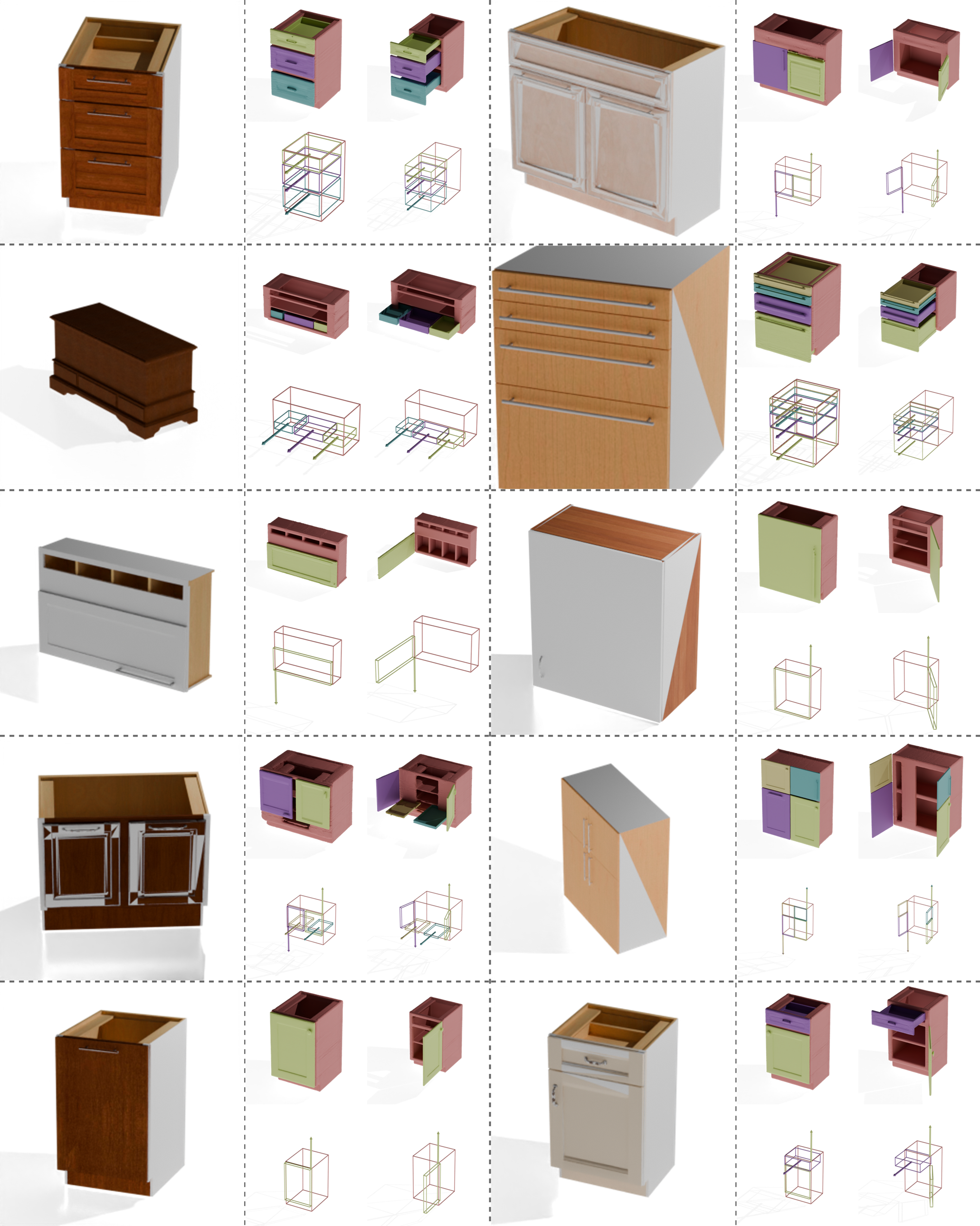}
    \caption{The figure presents 10 pairs of input images and articulated objects generated by ArtFormer. For each pair, the larger image on the left serves as the input image condition, while the right side illustrates the articulated object output, showcasing the predicted motion relationship with the joint in both the fully closed and fully open states.}
    \label{fig:image_cond_more}
\end{figure*}

\begin{figure*}[!htbp]
\centering
    \includegraphics[width=0.85\linewidth]{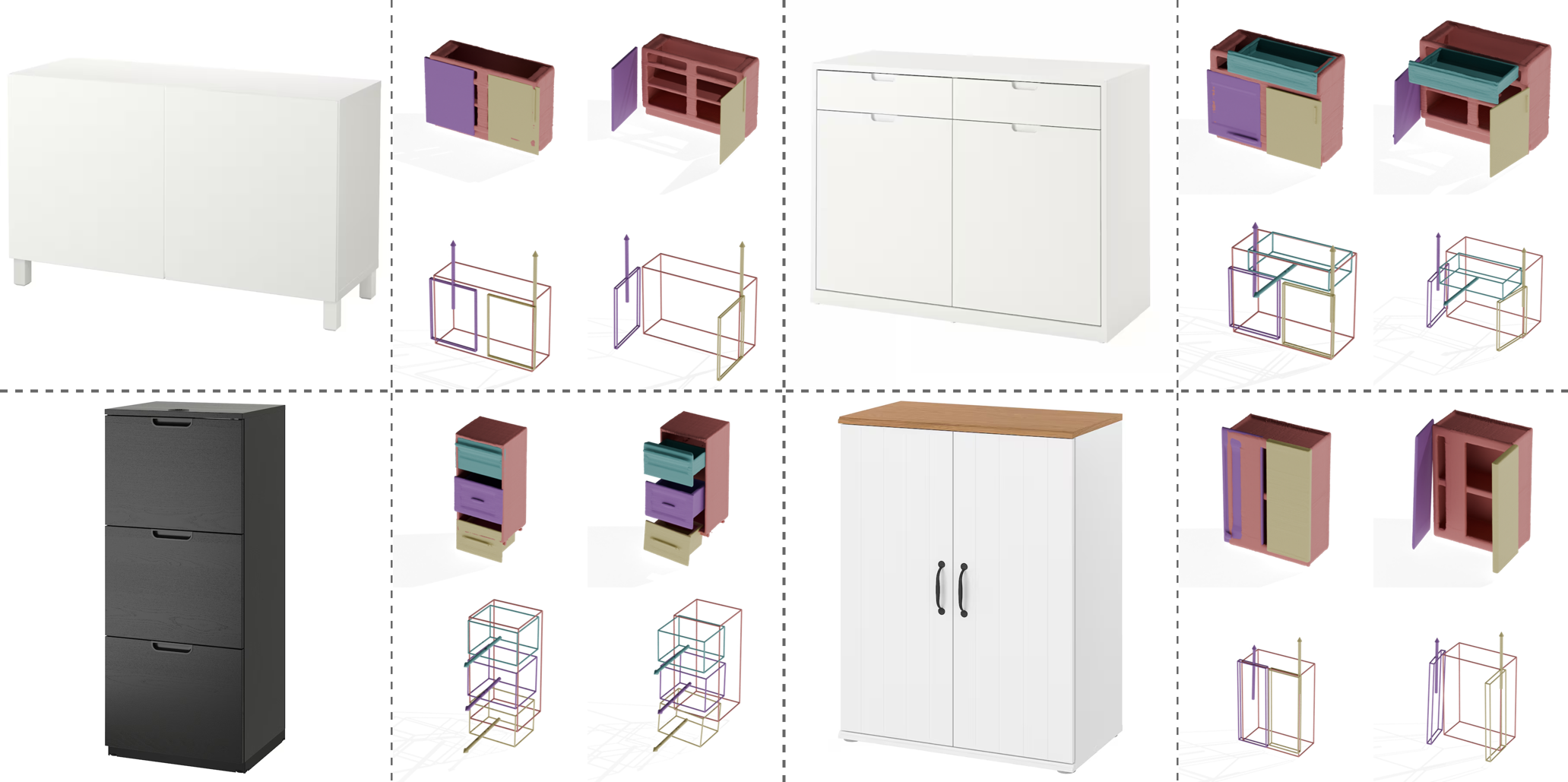}
    \caption{We present 4 pair of photographs for articulated objects from the real world (shown on the left side of each pair). Using our model, we generate these articulated object and make the visualization of them (shown on the right side of each pair).}
    \label{fig:real_image_cond}
\end{figure*}

\begin{table*}[h]
\caption{Text-Guided Object Alignment Results}
\centering
\begin{tabular}{@{}ccccc@{}}
\toprule
       & CAGE   & NAP-128 & NAP-768 & Ours   \\ \hline
CLIP-R@10$\uparrow$ & 0.1429 & 0.1648  & 0.1319  & \textbf{0.2198} \\ 
\bottomrule
\end{tabular}
\label{tab:clipr}
\end{table*}

\fi

\end{document}